%% file: main.tex
\documentclass[sigconf,nonacm]{acmart}
\AtBeginDocument{%
  }

\setcopyright{none}
\copyrightyear{2018}
\acmYear{2018}
\acmDOI{XXXXXXX.XXXXXXX}

\acmConference[Conference acronym 'XX]{Make sure to enter the correct
  conference title from your rights confirmation email}{June 03--05,
  2018}{Woodstock, NY}
\acmISBN{978-1-4503-XXXX-X/2018/06}

\renewcommand\footnotetextcopyrightpermission[1]{}

\usepackage{booktabs}
\usepackage{tabularx}
\usepackage[table]{xcolor}
\usepackage{multirow}
\usepackage{amsmath}
\usepackage{algorithm}
\usepackage{algpseudocode}

\algrenewcommand{\algorithmicrequire}{\textbf{Input:}}
\algrenewcommand{\algorithmicensure}{\textbf{Output:}}
\algrenewcommand{\alglinenumber}[1]{\scriptsize #1:}

\usepackage{booktabs,multirow}
\usepackage[table]{xcolor}

\definecolor{ablGain}{RGB}{0,128,0}
\definecolor{ablLoss}{RGB}{190,0,0}
\definecolor{ablNeutral}{RGB}{128,128,128}

\usepackage{pifont} 

\begin{document}

\title{TAPES: Task-Aligned Joint Predictive Pretraining for Transaction Event Sequences}


\settopmatter{printacmref=false,authorsperrow=1}

\author{%
  \texorpdfstring{%
    \begin{tabular}{@{}c@{}}
      \makebox[\linewidth][c]{\bfseries
        Ruobing Zuo\textsuperscript{1,2\textdagger},
        Weiyu Han\textsuperscript{2},
        Huilin He\textsuperscript{3,2\textdagger},
        Haoyu Jiang\textsuperscript{2},
        Ze Zhao\textsuperscript{2\textdagger},
        Jie Sheng\textsuperscript{2},
        }
      \\[0.2em]
      \textbf{%
        Xiao Tan\textsuperscript{2\ding{41}}, 
        Shuai Chen\textsuperscript{2},
        Tianyi Zhang\textsuperscript{2}%
      }
      \\[0.35em]
      \normalfont
      \textsuperscript{1}East China Normal University,
      \textsuperscript{2}Ant International,
      \textsuperscript{3}Tongji University
      \\[0.25em]
      \normalfont\ttfamily\fontsize{10}{12}\selectfont
        rbzuo@stu.ecnu.edu.cn, huilin3@tongji.edu.cn
        \\[0.1em]
        \normalfont\ttfamily\fontsize{10}{12}\selectfont
        \{hanweiyu.hwy,wuxian.jhy,zhaoze.zhao,shengjie.sheng,alex.tx,shuai.cs,zty113091\}@ant-intl.com
        \\[0.1em]
    \end{tabular}%
  }{Ruobing Zuo, Weiyu Han, Huilin He, Haoyu Jiang, Ze Zhao, Jie Sheng, Xiao Tan, Shuai Chen, Tianyi Zhang}%
}

\renewcommand{\shortauthors}{Zuo et al.}

\begin{abstract}
 Transaction event sequences support diverse prediction tasks in online commerce and digital financial services. 
 The same transaction records can yield prediction targets with different semantics, including values, changes, and relations to past events. These targets impose different constraints on history representations, but their predictability alone does not establish the downstream utility of a target set. Moreover, target-wise prediction losses do not explicitly model conditional co-occurrence among targets. 
 To address these challenges, 
we propose \textbf{TAPES}, a task-aligned joint predictive pretraining framework for transaction sequences. TAPES constructs a finite collection of candidate target sets from admissible field–operator combinations. 
It selects a target set according to the downstream utility of
representations learned through short joint pretraining trials
under matched budgets. 
It then reinitializes the encoder and pretrains it on the selected targets through a joint likelihood parameterized by a shared discrete latent variable, before downstream supervised adaptation.
 Experiments in a real-world transaction-risk setting and on multiple public financial benchmarks spanning diverse downstream tasks demonstrate that TAPES outperforms representative baselines. Ablation studies further validate the contributions of predictive target selection and joint modeling. 
\end{abstract}











\maketitle

\begingroup
\begin{NoHyper}
  \renewcommand{\thefootnote}{\textdagger}
  \footnotetext[0]{Work done during the internship at Ant International.}

  \renewcommand{\thefootnote}{\ding{41}}
  \footnotetext[0]{Corresponding author.}
\end{NoHyper}
\endgroup


\input{sections/Introduction}

\input{sections/Related_work}

\input{sections/Preliminary}

\input{sections/Method}


\input{sections/Experiments}

\input{sections/Conclusion}

\bibliographystyle{ACM-Reference-Format}
\bibliography{reference}

\appendix


\input{sections/Appedix}

\end{document}

%% file: sections/Introduction.tex
\section{Introduction}

E-commerce platforms and online payment services are integral to the Web economy. Purchases, payments, and related user interactions generate
continually growing transaction event sequences that support purchase prediction, user profiling, churn prediction, and risk
assessment~\cite{osin2025ebes,skalski2023foundation}. 
These histories capture heterogeneous event attributes and their temporal evolution, while different applications rely on different behavioral cues~\cite{bazarova2025learning}.
Transaction sequence pretraining provides a means of learning behavioral representations for downstream adaptation by deriving self-supervised signals from historical interactions~\cite{babaev2022coles,li2025panther}. 
\textit{Learning representations from rich transaction histories to support task-specific downstream decisions} is therefore an important problem in user behavior modeling on Web platforms. 


\begin{figure}[tbp]
  \centering
 \includegraphics[width=\linewidth]{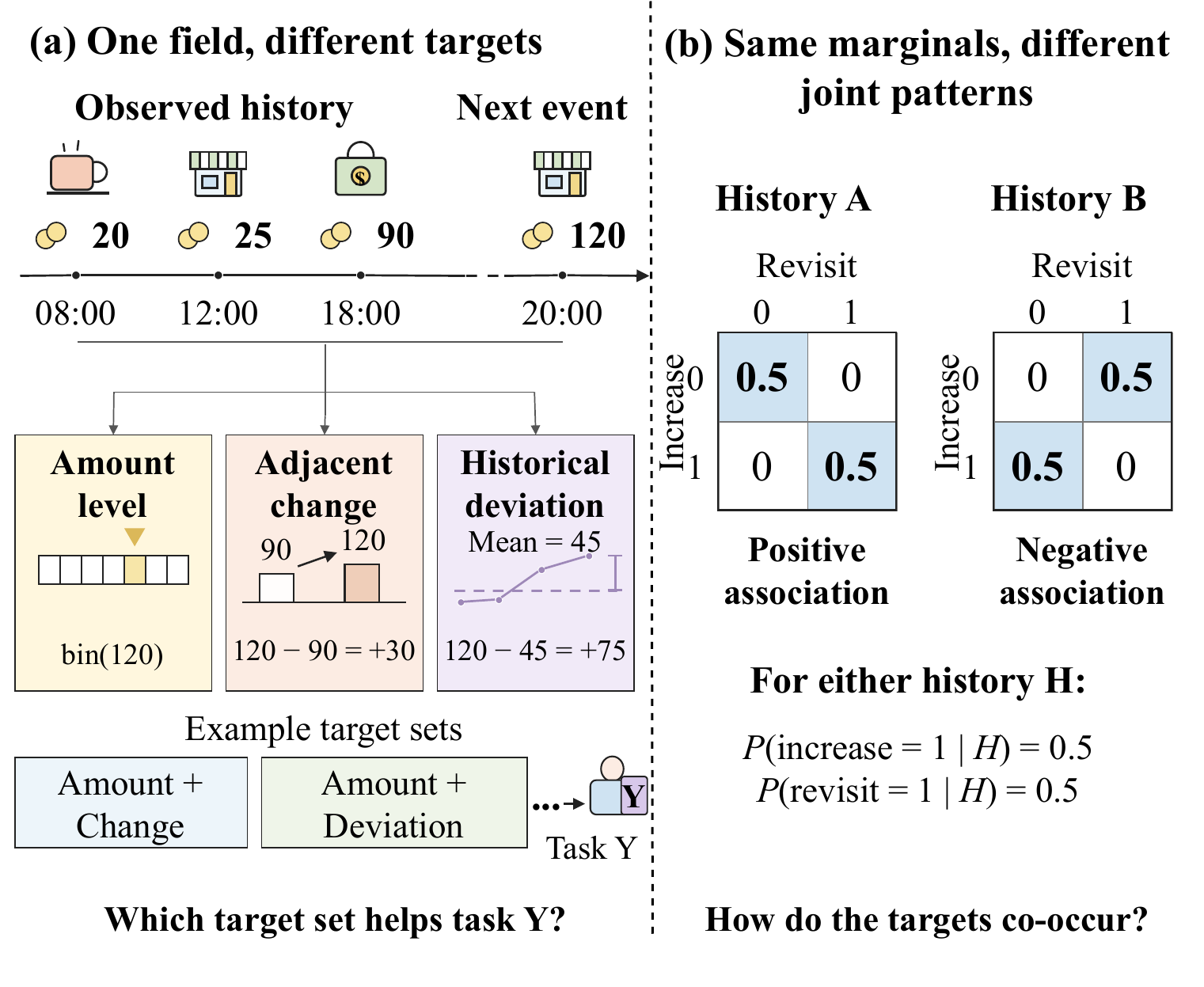}
 \caption{\textbf{Motivation for TAPES.}
(a) Different transformations of the same transaction field yield prediction targets with distinct semantics, illustrated by amount level, adjacent change, and historical deviation. 
(b) Two historical contexts share the same target-wise conditional
marginals but exhibit different joint outcome patterns,
motivating joint prediction.
Event values, target combinations, and distributions are illustrative;
$\operatorname{bin}$ denotes target-specific discretization.}
\label{fig:motivation}
\end{figure}

Existing studies have explored pretraining for transaction and behavioral sequences through different learning signals. 
{Contrastive approaches} learn behavioral features by aligning representations of different historical segments from the same entity~\cite{babaev2022coles}; 
{reconstruction- and prediction-based approaches} learn historical structure by recovering masked event attributes or predicting subsequent behavior~\cite{padhi2021tabular,transactiongpt}; 
{hybrid approaches} combine generative and contrastive supervision to exploit the complementary information they provide~\cite{moskvoretskii2024mlem}. These advances have promoted the learning of transferable representations from behavioral records and supported larger-scale modeling of payment behavior~\cite{transactiongpt}. However, pretraining tasks specify which properties of a history the encoder is trained to capture, while the value of these properties for a particular downstream task cannot be inferred directly from pretraining metrics. The same transaction history can provide multiple forms of predictive supervision, each imposing different learning constraints on the representation. 
Therefore, \textit{how to select and organize such supervision to better support downstream tasks} remains an important research question.
To investigate this question, we examine two interrelated design choices: \textit{which behavioral properties to predict} and \textit{how to predict them jointly}. 
\textbf{First}, a single transaction field supports different
predictive semantics, as illustrated in
Figure~\ref{fig:motivation}(a). 
For example, the next transaction amount can be described by its value range, its change from the preceding event, or its deviation from the mean of the observed history. These targets impose different requirements on history representations. 
Their predictability alone does not establish their downstream utility, and a target’s usefulness may depend on the other targets learned alongside it. 
This motivates evaluating complete target sets according to downstream representation utility. 
\textbf{Second}, predictive information may reside in conditional co-occurrence patterns among targets. 
As illustrated in Figure~\ref{fig:motivation}(b), two historical contexts may induce identical target-wise conditional marginals but different joint outcome distributions. 
Marginal objectives do not explicitly require distinguishing these joint patterns, whereas a joint likelihood can make them part of the learning signal. We therefore study \textbf{task-aligned target-set selection} and \textbf{joint prediction} of the selected targets.

Building on these considerations, we propose \textbf{TAPES}, a task-aligned joint predictive pretraining approach for transaction event sequences. TAPES comprises two stages. \textbf{Stage~1} constructs a finite collection of candidate target sets from admissible field--operator combinations. 
For each candidate set, a short joint pretraining trial produces
history representations whose downstream predictive performance
is evaluated using labels from within the training data.
The resulting evaluation scores guide target-set selection. 
\textbf{Stage~2} reinitializes the encoder while retaining the selected target definitions and fitted transformations. It then pretrains the encoder by optimizing a joint predictive likelihood with a shared discrete latent variable, before adapting the encoder to the downstream task through supervised fine-tuning. 
We compare TAPES with tree ensembles on aggregated features,
supervised sequence models, and self-supervised sequence pretraining
methods on five public benchmarks.
TAPES outperforms all evaluated baselines on the primary metric
in four benchmarks and achieves the best average rank of 1.20.
Cross-encoder comparisons and component ablations further support
its applicability across encoder architectures and the benefits
of both target-set selection and joint prediction.
Experiments on an industrial transaction-risk dataset containing
over 29 million transaction-level prediction examples further
examine the practical utility of pretrained history representations
for risk retrieval.

Our main contributions are summarized as follows:
\begin{itemize}
\item
    We incorporate the predictive semantics of transaction fields
    and the task utility of target sets into supervision design
    for transaction event sequences.
    Field--operator pairings define targets with distinct semantics,
    making explicit how the same event records can support
    different predictive learning objectives.

    \item
    We propose \textbf{TAPES}, a joint predictive pretraining approach
    that combines task-aligned target-set selection with conditional
    dependency modeling.
    TAPES selects predictive supervision according to downstream
    representation utility and learns history representations
    for downstream adaptation through joint prediction with
    a shared latent variable.


    \item
    We evaluate TAPES on five public benchmarks and a large-scale industrial transaction-risk dataset.
    Cross-encoder comparisons and component ablations support its applicability across multiple encoder architectures and the downstream contributions of both target-set selection and joint prediction.
\end{itemize}

%% file: sections/Related_work.tex
\section{Related Work}
\label{sec:related-work}

\subsection{Transaction and Behavioral Sequence Modeling}
\label{sec:rw-sequence-modeling}

\textit{Transaction and behavioral sequence modeling} aims to extract
task-relevant behavioral information from heterogeneous event
histories. Existing research encompasses feature-based
prediction~\cite{babaev2022coles} using gradient-boosted
trees~\cite{friedman2001gradient}, as well as supervised sequence
modeling~\cite{osin2025ebes} with recurrent~\cite{cho2014learning}
or attention-based encoders~\cite{vaswani2017attention}.

\textit{Pretraining} learns reusable representations through contrastive learning~\cite{babaev2022coles}, reconstruction and prediction~\cite{padhi2021tabular,skalski2023foundation},
or hybrid objectives~\cite{moskvoretskii2024mlem,castro2026abacus}.
Recent industrial models extend transaction pretraining with domain-specific representations and prediction objectives for large-scale payment behavior
modeling~\cite{li2025panther,transactiongpt,yeh2026treasure}. 
Other financial event models unify heterogeneous event sources into chronological sequences for next-event
pretraining~\cite{rusakov2026multimodal}.
Architectural work also introduces history tokens to accumulate prefix information during predictive pretraining for downstream sequence classification~\cite{karpukhin2026ht}. 
However, for configurations built around prespecified prediction
targets or statistics, supervision semantics do not automatically
adapt to downstream tasks.
Comparative studies~\cite{bazarova2025learning} also show that
methods differ in their relative strengths across local-behavior
and global-attribute tasks, highlighting the need to evaluate
supervision choices in relation to the target task.
TAPES makes this configuration choice explicit: it constructs
field--operator target sets while keeping the input history
unchanged and selects supervision based on the downstream utility
of representations learned by jointly predicting each set's targets.

\begin{figure*}[htbp]
  \centering
  \includegraphics[width=\linewidth]{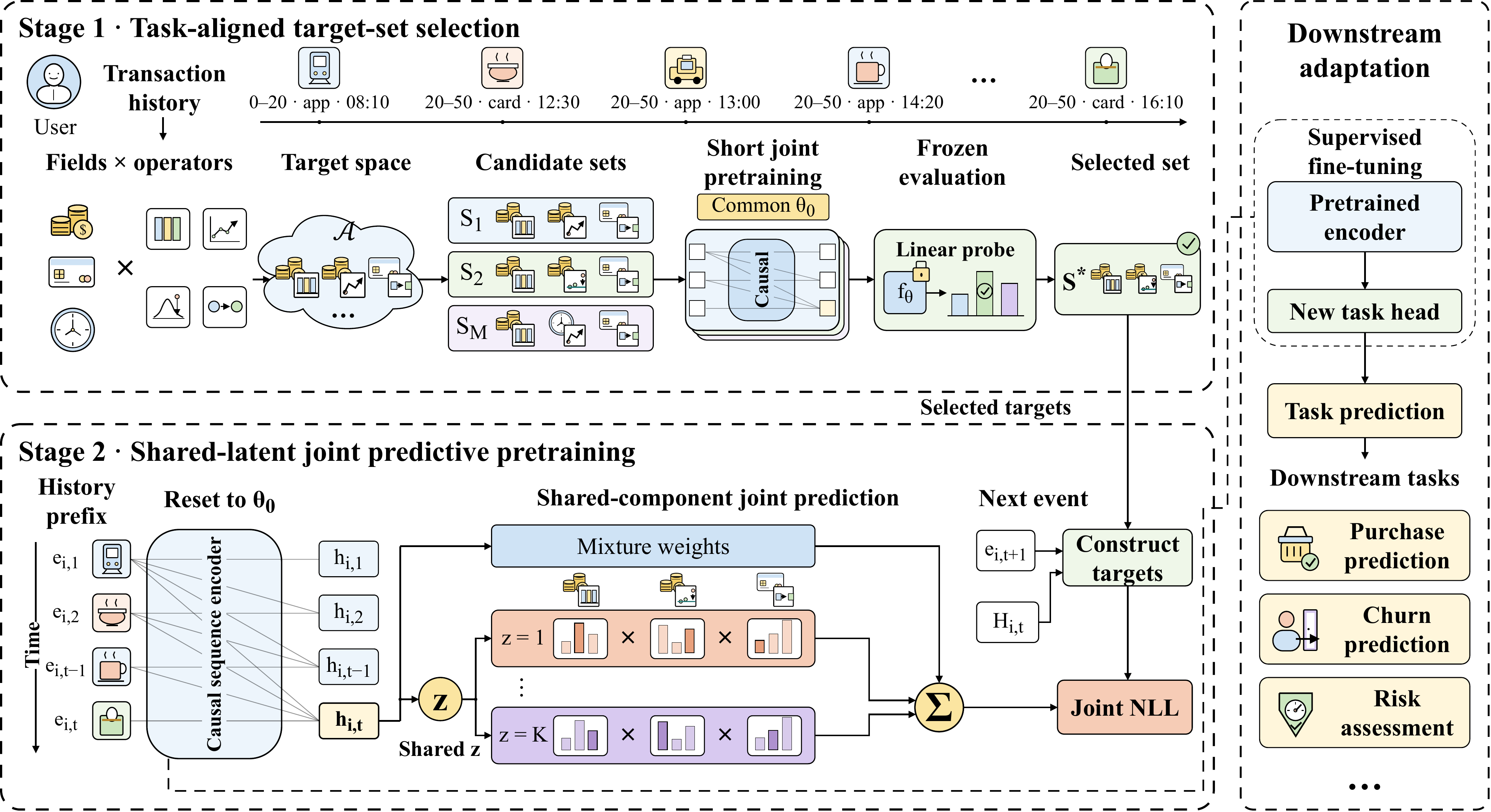}
 \caption{Overview of 
 Task-Aligned Joint Predictive Pretraining for Transaction Event Sequences 
 (TAPES). (1) Stage 1: Task-guided target-set selection. Field–operator combinations capture predictive semantics beyond raw field values, and short pretraining trials identify a target set useful for the downstream task. (2) Stage 2: Shared-latent joint pretraining. The selected targets are jointly modeled through a shared discrete latent variable to learn representations of transaction histories. Candidate sets and probabilities are illustrative.}
\label{fig:overview_tapes}
\end{figure*}

\subsection{Task-Guided Supervision Design}
\label{sec:rw-supervision-design}

The diversity of training signals has motivated research on
adapting supervision to downstream task requirements.
Task selection and mixing
methods~\cite{guo2019autosem,xue2020autoselect} adjust the
participation and relative contributions of auxiliary tasks;
auxiliary-loss learning and target
construction~\cite{navon2021auxilearn,dery2023aang} adapt
supervision content and composition;
task-grouping studies~\cite{standley2020tasks,fifty2021tag}
evaluate combinations through task-group training or affinity estimates obtained during optimization. 
Recent work also learns composite pretraining loss weights online by aligning pretraining gradients with a downstream objective, with evaluations including event-sequence
modeling~\cite{karpukhin2026grap}. 
These methods incorporate task combinations and downstream feedback
into supervision design through different search spaces and
evaluation protocols.
TAPES studies the configuration of field-derived predictive
semantics for transaction-history pretraining.
It compares predefined candidate sets in a field--operator semantic
space and uses the downstream validation performance of
representations obtained through short pretraining runs as a proxy
for the supervisory utility of each complete set.

\subsection{Joint Prediction and Target Dependencies}
\label{sec:rw-joint-prediction}

Joint prediction provides a framework for learning dependencies
among multiple outputs.
This perspective complements supervision selection: after
identifying useful targets, it remains necessary to determine
how they should be modeled jointly.
Shared-encoder multi-task learning~\cite{caruana1997multitask}
allows information transfer between targets, but summing marginal
prediction losses does not explicitly model their conditional
joint distribution.
Autoregressive conditional models~\cite{dembczynski2010pcc}
express output dependencies through chained conditional
factorizations.
The predictive value of exploiting dependencies depends on
the loss being optimized.
Shared-latent mixture models~\cite{li2016conditional} express
dependencies through mixtures of conditionally independent
distributions.

These models specify how to model dependencies within a given
output set, but do not by themselves select auxiliary-target
semantics relevant to downstream tasks.
TAPES combines dependency modeling with target-set selection,
using a shared discrete latent variable to define a
history-conditioned finite mixture of product categorical
distributions, thereby making joint configurations of the
selected targets part of the pretraining objective.

%% file: sections/Preliminary.tex
\section{Preliminaries}
\label{sec:preliminaries}

\textbf{Transaction Event Sequences.}
We study representation learning for transaction event sequences
to support downstream prediction tasks.
For entity \(i\), let
\[
H_{i,T_i}=(e_{i,1},\ldots,e_{i,T_i})
\]
denote the retained transaction history available at prediction
time, where \(T_i\) is the number of valid events and the events
are ordered chronologically.
Each event comprises the available categorical and numerical
attributes, temporal information, and field-level observation
indicators specifying whether individual fields are observed.
Examples of categorical and numerical attributes include event
type and transaction amount, respectively.
Individual fields may be missing, and events may occur at
irregular intervals.
We denote the observed prefix through event \(t\) by
\(H_{i,t}=(e_{i,1},\ldots,e_{i,t})\).

\textbf{Problem Formulation.}
For a given downstream task, each history is associated with
a label \(Y_i\in\mathcal Y\), where \(\mathcal Y\) is the
task-specific label space for binary, multiclass, or multilabel
prediction. 
Let \(f_\theta\) denote a causal sequence encoder incorporating
an event input module, and let \(g_\psi\) denote a task-specific
prediction function, including the output transformation
appropriate to the task. 
The history representation and downstream prediction are given by
\[
h_{i,t}=f_\theta(H_{i,t})\in\mathbb R^d,
\qquad
\widehat Y_i=g_\psi(h_{i,T_i}),
\]
where \(h_{i,T_i}\) is the representation at the last valid
event position, and \(\widehat Y_i\) denotes the predicted
probabilities. 
Causality requires that the representation at position \(t\)
depend only on the observed prefix \(H_{i,t}\), without access
to subsequent events.

\textbf{Task-Guided Predictive Pretraining.}
We consider a task-guided predictive pretraining setting in
which downstream labels within the training data guide the
selection of predictive target sets.
During event-prediction training, event-level targets are
constructed from an observed prefix \(H_{i,t}\) and its
corresponding next event \(e_{i,t+1}\), when available.
Although target construction uses the next event, the
representation \(h_{i,t}\) is computed exclusively from
\(H_{i,t}\).
The selected target definitions guide predictive pretraining,
after which the pretrained encoder is adapted to the target
downstream task through supervised fine-tuning.

%% file: sections/Method.tex

\section{Methodology}
\label{sec:tapes-method}

\subsection{Overview}

Predictive pretraining for transaction histories requires deciding which
event properties to predict and how to learn from them together. We introduce
\textbf{Task-Aligned Joint Predictive Pretraining for Transaction Event
Sequences (TAPES)}, which formulates supervision design as selection in a
structured field--operator target space. Stage~I selects complete target sets
according to the downstream utility of their jointly learned representations
(Section~\ref{sec:tapes-selection}). Stage~II trains the final encoder from
the common initialization using the selected supervision and the same joint
predictive loss (Section~\ref{sec:tapes-joint}), followed by downstream
supervised adaptation (Section~\ref{sec:tapes-transfer}).

\subsection{Stage 1: Task-Aligned Target-Set Selection}
\label{sec:tapes-selection}

\paragraph{Field--operator target space.}
A transaction field admits several predictive descriptions. An amount,
for example, can be described by its next value range, its adjacent change,
or its deviation from the historical level. Amount ranges emphasize
absolute scale, while prefix deviations emphasize departure from the
historical level. We make these predictive semantics explicit by defining
each target through an event field and a prediction operator.

Let $\mathcal F$ denote the available fields and $\mathcal O(f)$ the
admissible operators for field $f$. The target space and label construction
are
\begin{equation}
\begin{aligned}
\mathcal A
&=\{(f,o):f\in\mathcal F,\ o\in\mathcal O(f)\},\\
(y_{i,t,a},m_{i,t,a})
&=\Phi_a(H_{i,t},e_{i,t+1};\eta_a),
\qquad a\in\mathcal A.
\end{aligned}
\label{eq:tapes-target-space}
\end{equation}
Here, $y_{i,t,a}$ is a categorical prediction label,
$m_{i,t,a}\in\{0,1\}$ indicates whether the label can be validly constructed,
and $\eta_a$ contains the target transformation parameters. The next event
provides supervision, while the encoder receives the observed prefix
$H_{i,t}$. Selecting targets changes the output supervision while retaining
the available input fields.

Categorical operators describe state transitions and relations to previously
observed values; direct prediction is also available for admissible exact
low-cardinality vocabularies. Numerical operators describe value ranges,
adjacent differences, and deviations from the mean of valid prefix
observations. The resulting numerical quantities are discretized using
target-specific quantile boundaries, allowing heterogeneous targets to share
a categorical likelihood formulation. Target transformations are fitted on
predictive-pretraining data and held fixed across both stages.

\paragraph{Set-level downstream representation utility.}
A target's usefulness can depend on the other targets learned alongside it.
We therefore evaluate complete target sets through the representations they
jointly induce. Before candidate training, we construct a finite bank
$\mathcal C=(S_1,\ldots,S_M)$, where $S_j\subseteq\mathcal A$. Default
candidates are obtained by changing the field or operator of one target in
a common reference set while preserving set size. Dataset-specific banks
can also be prespecified from admissible targets.

For each candidate, we independently train an encoder and prediction module
from a common encoder initialization under the same training protocol,
using the shared-latent joint predictive loss adopted for final pretraining
(Section~\ref{sec:tapes-joint}).

We freeze each candidate encoder and fit a separate linear probe under a
common evaluation protocol. Predictive pretraining, probe fitting, and
candidate scoring use three disjoint subsets of the downstream training
set. The probe receives the last valid encoder state of each history and
remains fixed during scoring. We define the estimated set-level downstream
representation utility as
\begin{equation}
\begin{aligned}
\widehat U_J(S_j)
&=Q_{\mathrm{sel}}\!\left(
g_{\phi_j}\circ f_{\theta_j^{\mathrm I}}
\right),\\
S^*
&\in\operatorname*{arg\,max}_{S_j\in\mathcal C}
\widehat U_J(S_j).
\end{aligned}
\label{eq:tapes-selection}
\end{equation}
Here, $\theta_j^{\mathrm I}$ denotes the jointly pretrained candidate encoder
parameters, and $g_{\phi_j}$ is its fitted linear readout, including fixed
feature standardization. $Q_{\mathrm{sel}}$ evaluates the predictor against
downstream labels on the internal scoring subset using a prespecified
metric, with larger values indicating better performance.

\subsection{Stage 2: Shared-Latent Joint Predictive Pretraining}
\label{sec:tapes-joint}

\paragraph{Learning from joint outcomes.}
Single-target marginal distributions do not specify how target outcomes
co-occur. Different histories can therefore induce identical marginals
while exhibiting different joint patterns. To make target co-occurrence
part of the supervision, both pretraining stages use the following
shared-latent joint predictive model:
\begin{equation}
\begin{aligned}
p_\omega(\mathbf y_{i,t,S}\mid h_{i,t})
&=\sum_{k=1}^{K}\pi_k(h_{i,t})
\prod_{a\in S}q_{a,k}(y_{i,t,a}\mid h_{i,t}),\\
h_{i,t}&=f_\theta(H_{i,t}).
\end{aligned}
\label{eq:tapes-joint}
\end{equation}
Here, $K$ is the number of components, $\pi_k(h_{i,t})$ is the probability
of component $k$, $q_{a,k}$ is its categorical distribution for target $a$,
and $\omega$ denotes the prediction-module parameters. Targets factorize
conditional on the history representation and component; marginalizing
the shared component allows dependencies among their outcomes.

Within each component, target probabilities are multiplied before the
components are mixed. The joint likelihood thus evaluates how well the
same component explains multiple observed outcomes together, coupling
their contributions to the encoder update.

When only part of the target vector is available, we marginalize unavailable
coordinates by restricting the product in Equation~\eqref{eq:tapes-joint}
to valid targets. The joint predictive loss is
\begin{equation}
\mathcal L_J(S)
=-\frac{1}{|S|\,|\mathcal P_S|}
\sum_{(i,t)\in\mathcal P_S}
\log p_\omega\!\left(
\mathbf y_{i,t,O_{i,t}(S)}\mid h_{i,t}
\right),
\label{eq:tapes-loss}
\end{equation}
where $O_{i,t}(S)=\{a\in S:m_{i,t,a}=1\}$, and $\mathcal P_S$ is the
nonempty set of adjacent event pairs with at least one valid target in the
training batch. The loss averages over valid event pairs and scales by the
fixed target-set size $|S|$. Missingness encoded as an explicit target
category remains valid supervision.

\paragraph{Learning with the selected supervision.}
Stage~II retains $S^*$ and its fitted target transformations. The encoder
is reset to the common initialization, and the prediction module and
optimizer are reinitialized. We then minimize $\mathcal L_J(S^*)$ on the
predictive-pretraining data to learn the final history encoder.





\subsection{Downstream Transfer and Adaptation}
\label{sec:tapes-transfer}

\paragraph{Parameter transfer.}
For standard sequence classification, we retain the event input
module and history encoder learned in Stage~II, initializing
their parameters as
\(\theta_{\mathrm{init}}=\theta^{\mathrm{II}}\),
and discard the joint prediction module.
A newly initialized affine head predicts from the last valid
event state:
\begin{equation}
\begin{aligned}
z_i &= W_\psi h_{i,T_i}+b_\psi,\\
\widehat Y_i &= g_\psi(h_{i,T_i})
             = \rho_{\mathrm{task}}(z_i).
\end{aligned}
\label{eq:tapes-downstream-prediction}
\end{equation}
Here, \(z_i\) denotes the logits, and \(\widehat Y_i\) denotes
the predicted probabilities.
The prediction function \(g_\psi\) combines the affine head
with \(\rho_{\mathrm{task}}\), which is sigmoid for binary
and multilabel prediction and softmax for multiclass prediction.
Padding positions are excluded when identifying the last valid
state.
The available input fields and original preprocessing are
retained; the Stage~I probe and its fixed
representation-standardization statistics are not reused.

\paragraph{Supervised fine-tuning.}
With a newly initialized optimizer, we jointly update the
transferred modules and the task head by minimizing
\begin{equation}
\mathcal L_{\mathrm{task}}(\theta,\psi)
=
\frac{1}{|\mathcal D_{\mathrm{fit}}|}
\sum_{i\in\mathcal D_{\mathrm{fit}}}
\ell_{\mathrm{sup}}(z_i,Y_i),
\label{eq:tapes-adaptation}
\end{equation}
where \(\ell_{\mathrm{sup}}\) is the task-specific supervised
loss computed from logits.
Binary classification and multilabel prediction with fully
observed labels use binary cross-entropy, while multiclass
prediction uses multiclass cross-entropy. 
The joint predictive loss \(\mathcal L_J\) is no longer optimized.
The fine-tuning set \(\mathcal D_{\mathrm{fit}}\) comprises all
three internal training subsets and their downstream labels,
excluding the dedicated early-stopping validation and
development evaluation sets.
Stage~2 event pretraining remains restricted to its original
predictive-pretraining subset.

Target-set selection and predictive pretraining are performed
separately for each downstream task.
For tasks with established prediction architectures, we transfer
the corresponding pretrained modules while retaining the
task-specific prediction and fusion structures and supervised
objectives.
Appendix~\ref{app:downstream} provides additional details
on downstream adaptation and evaluation.

%% file: sections/Experiments.tex
\providecommand{\pending}{\textit{Pending}}
\providecommand{\tbd}{\textit{TBD}}

\section{Experiments}
\label{sec:experiments}
We compare TAPES with baseline methods on five public benchmarks and examine its performance across encoder configurations and fine-tuning label budgets. Component ablations assess target-set selection and joint prediction, while additional studies examine industrial applications and the pretraining budget. 


\subsection{Experimental Setup}
\label{sec:exp-setup}

Our experiments cover multiple downstream tasks on public transaction and
behavioral sequences, together with industrial transaction-risk assessment,
to examine predictive performance across label spaces and data scales.

\begin{table*}[t]
\centering
\caption{Development results for the evaluated configurations on five public benchmarks (\%).
PANTHER uses the SPRM adaptation specified in Appendix~\ref{app:baselines}.
Bold and underline indicate the \textbf{best} and \underline{second-best} scores in each column.
Avg.~Rank is the mean rank over the five primary metrics. 
}
\label{tab:main-comparison}

\begingroup
\normalsize
\setlength{\tabcolsep}{1.7pt}
\renewcommand{\arraystretch}{1.08}

\begin{tabular}{@{}l*{12}{c}@{}}
\toprule
\multirow{2}{*}{Method}
& \multicolumn{2}{c}{\shortstack{AlfaBattle}}
& \multicolumn{2}{c}{\shortstack{Churn}}
& \multicolumn{2}{c}{Taobao}
& \multicolumn{2}{c}{MBD-mini}
& \multicolumn{3}{c}{Retail X5}
& \multirow{2}{*}{\shortstack{Avg.Rank$\downarrow$}}
\\

\cmidrule(lr){2-3}
\cmidrule(lr){4-5}
\cmidrule(lr){6-7}
\cmidrule(lr){8-9}
\cmidrule(lr){10-12}

& AUROC$\uparrow$ & AP$\uparrow$
& AUROC$\uparrow$ & AP$\uparrow$
& AUROC$\uparrow$ & AP$\uparrow$
& mAUROC$\uparrow$ & mAP$\uparrow$
& \shortstack{Macro F1}$\uparrow$
& Acc.$\uparrow$
& AUROC$\uparrow$
&
\\

\midrule
\multicolumn{13}{l}{\textit{Tree ensembles on aggregated features}} \\%

CatBoost
& 76.64 & \underline{12.80}
& 81.60 & 84.39
& \underline{70.66} & \underline{75.61}
& 77.60 & 1.81
& 49.00 & 51.44 & 77.17
& 5.40
\\

XGBoost
& \underline{76.95} & \textbf{12.87}
& 81.82 & 84.05
& 70.65 & 75.30
& \underline{79.33} & 1.75
& 49.86 & 51.85 & 77.53
& \underline{4.00}
\\

\midrule
\multicolumn{13}{l}{\textit{Supervised sequence models}} \\

GRU
& 74.33 & 10.27
& 81.48 & 84.52
& 68.08 & 72.85
& 78.05 & 1.31
& \underline{53.79} & \textbf{54.80} & \textbf{80.17}
& 6.00
\\

LSTM
& 74.78 & 10.46
& 80.90 & 83.58
& 67.15 & 71.93
& 77.15 & 1.49
& 52.97 & 54.12 & 79.50
& 7.10
\\

Transformer
& 75.57 & 11.01
& 80.90 & 83.93
& 69.85 & 74.74
& 79.27 & \textbf{2.01}
& 53.58 & 54.30 & 79.64
& 4.90
\\

Transformer (RoPE)
& 76.07 & 11.26
& 81.73 & 83.96
& 70.11 & 74.83
& 79.17 & 1.60
& 52.32 & 53.43 & 79.47
& 4.40
\\

\midrule
\multicolumn{13}{l}{\textit{Self-supervised pretraining}} \\

CoLES
& 71.30 & 8.39
& \textbf{83.81} & 84.67
& 68.12 & 74.22
& 73.61 & 0.89
& 51.97 & 53.50 & 79.12
& 7.40
\\

TabBERT
& 72.01 & 8.71
& 76.59 & 80.74
& 68.14 & 73.06
& 76.50 & 1.19
& 51.65 & 53.21 & 78.58
& 8.40
\\

MLEM
& 71.59 & 8.35
& 77.85 & 80.93
& 68.72 & 74.16
& 76.09 & 0.95
& 52.21 & 53.46 & 79.07
& 8.00
\\

PANTHER
& 70.48 & 8.48
& 82.68 & \textbf{85.92}
& 63.80 & 67.76
& 76.04 & 1.20
& 30.63 & 35.16 & 60.94
& 9.20
\\

\midrule
\rowcolor[RGB]{245,242,252}
\textbf{TAPES(Ours)}
& \textbf{77.38} & 12.52
& \underline{82.99} & \underline{85.03}
& \textbf{71.03} & \textbf{76.16}
& \textbf{79.95} & \underline{1.91}
& \textbf{54.08} & \underline{54.60} & \underline{79.90}
& \textbf{1.20}
\\

\bottomrule
\end{tabular}

\endgroup
\end{table*}

\paragraph{Datasets.}
We evaluate TAPES on five public benchmarks for transaction and
user-behavior sequences:
Alpha\footnote{\url{https://boosters.pro/championship/alfabattle2/overview}},
Retail X5\footnote{\url{https://ods.ai/competitions/x5-retailhero-uplift-modeling/data}},
MBD-mini~\cite{mollaev2025mbd}\footnote{\url{https://huggingface.co/datasets/ai-lab/MBD-mini}},
Rosbank Churn\footnote{\url{https://boosters.pro/championship/rosbank1/}},
and Taobao\footnote{\url{https://tianchi.aliyun.com/dataset/46}}.
These benchmarks cover credit default prediction, four-class age
classification, next-month purchase prediction for four banking
products, customer churn prediction, and purchase prediction over
the following seven days, respectively. 
We additionally evaluate TransRisk, a private industrial dataset for binary
transaction-risk prediction containing over 29 million transaction-level
prediction examples across its training and validation partitions.
Each example pairs a current transaction with the corresponding user history.
Dataset sources, statistics, and task definitions are provided in
Appendix~\ref{app:datasets}. 

\paragraph{Baselines.}
We compare three categories of baselines to examine different ways of
learning predictive information from event histories.
(i) \emph{Tree ensembles on aggregated features} include
CatBoost~\cite{NEURIPS2018_catboost} and XGBoost~\cite{xgboost2016}.
They learn supervised predictors from statistical summaries of
event histories, providing references based on aggregated behavioral information.
(ii) \emph{Supervised sequence encoders} include GRU~\cite{cho2014learning},
LSTM~\cite{hochreiter1997lstm}, a causal Transformer~\cite{vaswani2017attention},
and its RoPE configuration. These models learn the sequence encoder
and task head directly from downstream labels.
In the encoder study, supervised training of each encoder provides
the corresponding Scratch reference for TAPES.
(iii) \emph{Self-supervised sequence pretraining} includes
CoLES~\cite{babaev2022coles}, which learns from contrasting subsequence views;
TabBERT~\cite{padhi2021tabular}, which reconstructs masked fields;
and MLEM~\cite{moskvoretskii2024mlem}, which aligns generative and
contrastive representations.
We also evaluate PANTHER-SPRM, an adaptation of PANTHER's public
sequential pretraining component~\cite{li2025panther}.
These methods use project-adapted implementations for the event inputs
and downstream tasks considered here.
Implementation details and model configurations are given in
Appendices~\ref{app:baselines} and~\ref{app:models-training}.

\paragraph{Models and training.} 
We evaluate recurrent and Transformer encoders, including GRU,
LSTM, a standard causal Transformer, and Transformer configurations
with rotary positional embeddings.
The main comparison uses method- and dataset-specific encoder
configurations.
In the cross-encoder study, each TAPES configuration is compared
with a supervised reference using the same encoder architecture.
Detailed encoder configurations and training budgets are provided
in Appendix~\ref{app:models-training}. 

We compare against aggregated-feature CatBoost and XGBoost models,
supervised sequence encoders, and existing self-supervised sequence
methods. 

\paragraph{Evaluation.} 
For each public benchmark, we select the fine-tuning checkpoint
according to the primary task metric on a dedicated validation set.
We then report the selected checkpoint's performance on a separate
development evaluation set. 
The primary metric is AUROC for 
AlfaBattle, Rosbank Churn, and Taobao, macro-AUROC for MBD-mini, and Macro-F1
for Retail X5. We also report AP and dataset-specific supplementary metrics.
For TransRisk, we evaluate ranking quality using AUROC and risk retrieval
using Recall@top1\% and AddCB recall at a 0.5\% intervention budget.
Table~\ref{tab:main-comparison} summarizes the evaluated model configurations.
Encoder configurations, training schedules, and detailed evaluation settings 
are provided in Appendices~\ref{app:models-training} and 
\ref{app:evaluation-details}. 


\subsection{Main Results}
\label{sec:exp-overall}

We first compare TAPES with the three baseline categories on the five
public benchmarks, then examine performance across encoder choices,
fine-tuning label budgets, and industrial risk-retrieval settings.

\begin{table*}[t]
\centering
\caption{
Development performance across encoder configurations (\%).
Parentheses show relative percentage changes from the displayed supervised score. 
Retail uses Macro-F1, MBD-mini uses macro-AUROC, and the remaining datasets use AUROC. 
$d$ and $L$ denote hidden dimension and depth.
Bold marks the higher score within each two-row group.
}
\label{tab:backbone-comparison}

\begingroup
\normalsize
\setlength{\tabcolsep}{3pt}
\renewcommand{\arraystretch}{1.08}

\resizebox{0.8\textwidth}{!}{%
\small
\begin{tabular}{lccccccc}
\toprule
Backbone & $d$ & $L$
& \shortstack{AlfaBattle}
& Retail X5 & MBD
& \shortstack{Churn}
& Taobao \\

\midrule
GRU & 512 & 2
& 74.59 & \textbf{54.34} & \textbf{78.05}
& 81.48 & 68.08 \\

\rowcolor[RGB]{245,242,252}
\quad \textbf{+ TAPES(Ours)} & &
& \textbf{76.05}\,{\footnotesize(+1.96\%)}
& 53.85\,{\footnotesize(-0.90\%)}
& 75.93\,{\footnotesize(-2.72\%)}
& \textbf{81.56}\,{\footnotesize(+0.10\%)}
& \textbf{71.03}\,{\footnotesize(+4.33\%)} \\

\midrule
LSTM & 512 & 2
& 74.78 & 52.97 & 77.15 & 80.90 & 67.15 \\

\rowcolor[RGB]{245,242,252}
\quad \textbf{+ TAPES(Ours)} & &
& \textbf{75.51}\,{\footnotesize(+0.98\%)}
& \textbf{53.37}\,{\footnotesize(+0.76\%)}
& \textbf{77.41}\,{\footnotesize(+0.34\%)}
& \textbf{81.78}\,{\footnotesize(+1.09\%)}
& \textbf{70.36}\,{\footnotesize(+4.78\%)} \\

\midrule
Transformer & 256 & 4
& 75.57 & 53.60 & 79.27 & 80.90 & 69.85 \\

\rowcolor[RGB]{245,242,252}
\quad \textbf{+ TAPES(Ours)} & &
& \textbf{78.09}\,{\footnotesize(+3.33\%)}
& \textbf{54.30}\,{\footnotesize(+1.31\%)}
& \textbf{79.95}\,{\footnotesize(+0.86\%)}
& \textbf{82.99}\,{\footnotesize(+2.58\%)}
& \textbf{70.55}\,{\footnotesize(+1.00\%)} \\

\midrule
Transformer (RoPE) & 256 & 4
& 75.58 & 52.15 & 78.33 & 81.73 & 68.83 \\

\rowcolor[RGB]{245,242,252}
\quad \textbf{+ TAPES(Ours)} & &
& \textbf{77.26}\,{\footnotesize(+2.22\%)}
& \textbf{53.08}\,{\footnotesize(+1.78\%)}
& \textbf{79.23}\,{\footnotesize(+1.15\%)}
& \textbf{82.08}\,{\footnotesize(+0.43\%)}
& \textbf{69.66}\,{\footnotesize(+1.21\%)} \\

\midrule
Transformer (RoPE) & 1024 & 8
& 76.07 & 52.32 & \textbf{79.17} & 80.97 & 70.11 \\

\rowcolor[RGB]{245,242,252}
\quad \textbf{+ TAPES(Ours)} & &
& \textbf{77.38}\,{\footnotesize(+1.72\%)}
& \textbf{52.70}\,{\footnotesize(+0.73\%)}
& 79.04\,{\footnotesize(-0.16\%)}
& \textbf{82.21}\,{\footnotesize(+1.53\%)}
& \textbf{70.85}\,{\footnotesize(+1.06\%)} \\

\bottomrule
\end{tabular}}
\endgroup
\end{table*}

\begin{figure*}[h]
  \centering
  \includegraphics[width=0.95\linewidth]{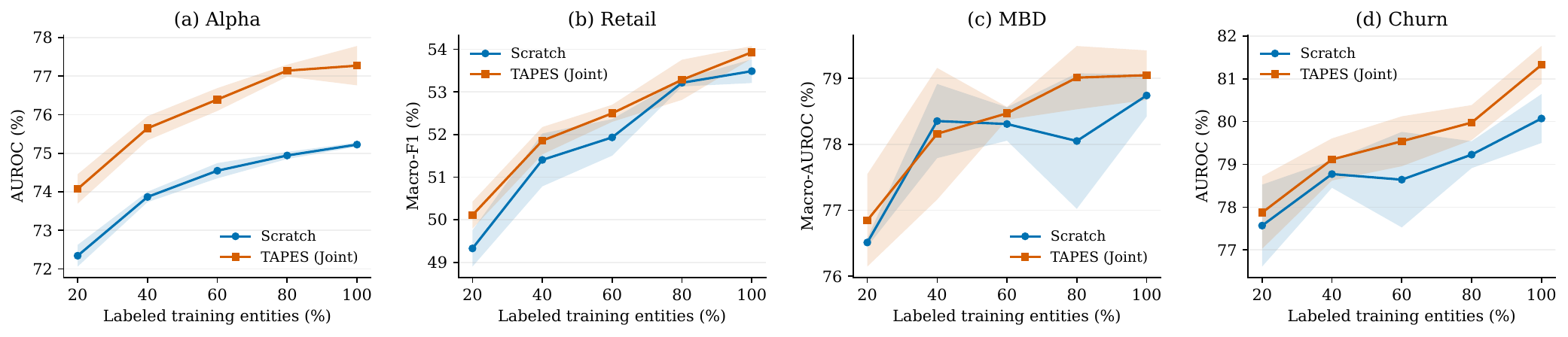}
    \caption{Development performance at different labeled-data fractions for 
downstream fine-tuning on 
AlfaBattle, Retail X5, MBD-mini, and Rosbank Churn. 
Curves show means over 
fine-tuning seeds; shaded regions indicate 
$\pm$ one sample standard deviation. 
}
\label{fig:label-efficiency}
\end{figure*}

\providecolor{tapeslight}{HTML}{F1EBF8}
\begin{table}[t]
\centering
\caption{Validation results on TransRisk after 25k fine-tuning updates. 
Bold indicates the best value.  
}
\label{tab:industrial}
\small
\setlength{\tabcolsep}{5pt}
\begin{tabular}{lrrr}
\toprule
Method & AUROC (\%) & \shortstack{Passed-CB\\AP (\%)} & \shortstack{Relative AddCB\\change (\%)} \\
\midrule
Scratch & 90.88 & \textbf{5.80} & 0.00 \\ 
CoLES & 89.66 & 4.35 & -0.93 \\
TabBERT & 90.83 & 4.95 & -0.93 \\
MLEM & 90.13 & 5.47 & \textbf{+0.28} \\
\midrule
 \rowcolor{tapeslight}
\textbf{TAPES (Ours)} & \textbf{92.04} & 4.48 & \textbf{+0.30} \\
\bottomrule
\end{tabular}
\end{table}

\begin{figure}[htbp]
  \centering
 \includegraphics[width=\linewidth]{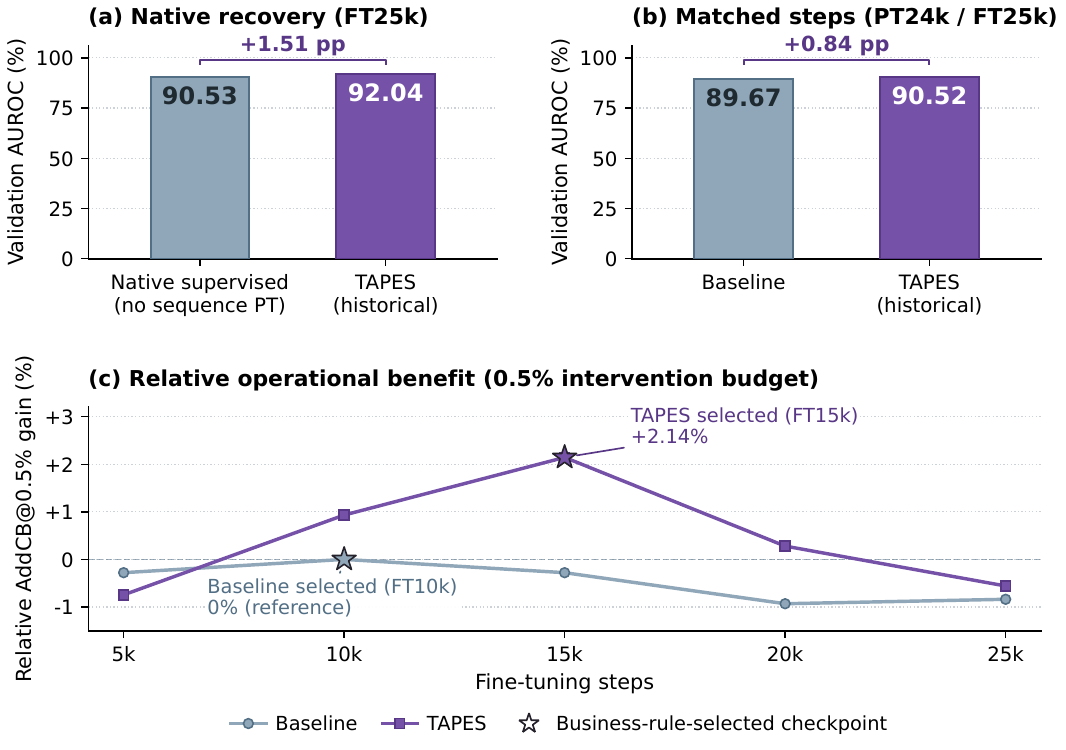}
\caption{Industrial fine-tuning evaluations on TransRisk.
(a) AUROC of TAPES and the supervised control without
history pretraining after 25k fine-tuning updates. 
(b) Comparison with baseline using 24k pretraining updates
and 25k fine-tuning updates. 
(c) Relative AddCB recall change at a 0.5\% intervention budget, 
with the business baseline selected at 10k updates as reference. 
Stars identify checkpoints selected by the validation business criterion. 
}
\label{fig:industrial-finetuning}
\end{figure}

\begin{figure}[htbp]
  \centering
 \includegraphics[width=\linewidth]{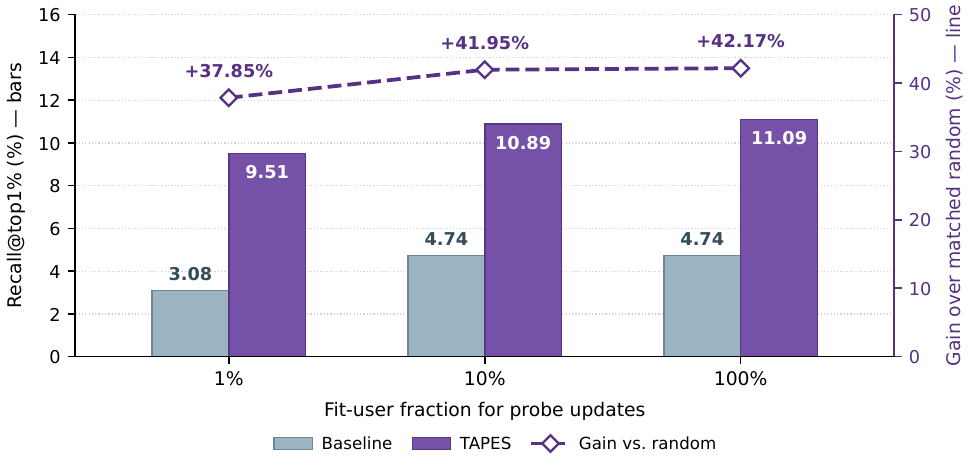}
\caption{Frozen-representation evaluation on TransRisk.
Bars show Recall@top1\% for TAPES and the business baseline (left axis).
The dashed curve shows TAPES's relative recall gain over an
architecture-matched random encoder (right axis).
}
\label{fig:industrial-frozen}
\end{figure}

\textbf{TAPES achieves the best primary score on four of the five public benchmarks.}
Table~\ref{tab:main-comparison} summarizes the public benchmark results
and averages method ranks over the designated primary metrics.
TAPES exceeds the strongest reported baseline by 0.43, 0.37, 0.62, and 0.29 percentage points on AlfaBattle,
Taobao, MBD-mini, and Retail X5, respectively.
It ranks second on Rosbank Churn, where CoLES obtains 83.81 AUROC
compared with 82.99 for TAPES.
TAPES has the lowest average rank of 1.20 across the five primary metrics.
Supplementary metrics highlight additional strengths of the baselines:
XGBoost obtains the highest AP on AlfaBattle, the Transformer the highest
macro-AP on MBD-mini, and GRU the highest accuracy and AUROC on Retail X5. 
Overall, these results support the utility of task-aligned predictive pretraining across diverse transaction and behavioral prediction tasks.

\textbf{TAPES improves performance across multiple encoder configurations.}
To examine its compatibility with different encoder architectures,
we apply TAPES to GRU, LSTM, a standard Transformer, and two RoPE
configurations, and compare each with its directly supervised counterpart.
Table~\ref{tab:backbone-comparison} shows improvements in 22 of 25
encoder--dataset combinations.
All five tasks improve with LSTM, the standard Transformer, and the
smaller RoPE configuration. The exceptions are GRU on Retail X5 and MBD-mini,
and the larger RoPE configuration on MBD-mini.
The encoder-specific configurations, seeds, and pretraining schedules
are summarized in Appendix~\ref{app:models-training}.

\textbf{TAPES retains higher mean scores at most fine-tuning label fractions.} 
To assess the value of pretrained representations as downstream labels
become scarcer, 
we compare TAPES with Scratch on four public benchmarks using
20\%, 40\%, 60\%, 80\%, and 100\% of the downstream training labels.
For each benchmark, we reuse the same TAPES pretrained checkpoint
and Stage~1 selection across label budgets.
Figure~3 reports mean scores over three fine-tuning seeds.
TAPES achieves higher scores in 19 of the 20 settings and can
outperform Scratch with fewer labeled examples. 
For example, on AlfaBattle, TAPES achieves an AUROC of 75.65 with only 40\% of the labels, surpassing the 75.22 achieved by Scratch using all labels. 

\textbf{Fine-tuning improves industrial ranking and budget- constrained risk retrieval.}
To assess the downstream value of pretrained history representations,
we examine industrial fine-tuning under fixed training endpoints and
business-based checkpoint selection.  
Figure~\ref{fig:industrial-finetuning}(a) 
evaluates TAPES in the industrial risk model
after 25,000 fine-tuning updates.
TAPES achieves 92.04\% AUROC, outperforming the supervised
control without history pretraining by 1.51 percentage
points. 
With 24k pretraining and 25k fine-tuning updates in panel (b),
TAPES achieves 90.52\% AUROC versus 89.67\% for the business baseline,
corresponding to a 0.84-point gain calculated from the original scores.
Panel (c) evaluates risk capture at a 0.5\% intervention budget.
The validation business criterion selects TAPES at 15k updates and the
business baseline at 10k; TAPES increases AddCB recall by 2.14\%
relative to this selected baseline.
These results show gains in both ranking quality and risk retrieval
at the corresponding evaluation settings.

\newcommand{\abldiff}[1]{%
  \,{\scriptsize
  \ifdim #1pt>0pt
    \color{ablGain}%
  \else
    \ifdim #1pt<0pt
      \color{ablLoss}%
    \else
      \color{ablNeutral}%
    \fi
  \fi
  (#1)}%
}

\begin{table*}[htbp]
\centering
\caption{
Component ablation under full fine-tuning.
$S$: target-set selection; $J$: joint likelihood.
All metrics within each configuration--dataset pair
come from the same checkpoint.
Parenthesized values indicate changes in percentage points
relative to w/o Pretraining.
Green and red indicate increases and decreases, respectively.
AP denotes average precision;
Retail AUROC is macro one-vs-rest AUROC.
Bold marks the best scores at the reported precision.
}
\label{tab:component-ablation}

\begingroup
\normalsize
\setlength{\tabcolsep}{3pt}
\renewcommand{\arraystretch}{1.12}

\begin{tabular}{@{}lcc*{5}{r}@{}}
\toprule
\multirow{2}{*}{Ablation Variant}
& \multirow{2}{*}{$S$}
& \multirow{2}{*}{$J$}
& \multicolumn{2}{c}{AlfaBattle}
& \multicolumn{3}{c}{Retail X5} \\
\cmidrule(lr){4-5}\cmidrule(l){6-8}
& & &
AUROC$\uparrow$ & AP$\uparrow$
& Macro-F1$\uparrow$ & Acc.$\uparrow$ & AUROC$\uparrow$ \\

\midrule
w/o Pretraining
& -- & --
& 75.25
& 10.80
& 53.18
& 54.17
& 79.66 \\

w/o Both
& $\times$ & $\times$
& 76.98\abldiff{+1.73}
& 12.01\abldiff{+1.21}
& 53.32\abldiff{+0.14}
& 54.15\abldiff{-0.02}
& 79.78\abldiff{+0.12} \\

w/o Selection
& $\times$ & $\checkmark$
& 77.09\abldiff{+1.84}
& 12.18\abldiff{+1.38}
& 53.40\abldiff{+0.22}
& 54.25\abldiff{+0.08}
& 79.82\abldiff{+0.16} \\

w/o Joint Prediction
& $\checkmark$ & $\times$
& 77.12\abldiff{+1.87}
& 12.35\abldiff{+1.55}
& 53.79\abldiff{+0.61}
& 54.47\abldiff{+0.30}
& \textbf{79.90}\abldiff{+0.24} \\

\midrule
\rowcolor[RGB]{245,242,252}
\textbf{TAPES (Ours)}
& $\checkmark$ & $\checkmark$
& \textbf{77.38}\abldiff{+2.13}
& \textbf{12.52}\abldiff{+1.72}
& \textbf{54.08}\abldiff{+0.90}
& \textbf{54.60}\abldiff{+0.43}
& \textbf{79.90}\abldiff{+0.24} \\

\bottomrule
\end{tabular}
\endgroup
\end{table*}

\textbf{Frozen industrial representations improve top-ranked risk retrieval.}
To examine the utility of history representations for industrial risk
prediction, we freeze the sequence encoder on TransRisk and fit affine
probes with different fractions of training users.
Figure~\ref{fig:industrial-frozen} reports the resulting top-ranked risk retrieval.
TAPES obtains Recall@top1\% of 9.51\%, 10.89\%, and 11.09\%
using 1\%, 10\%, and 100\% of training users for probe-gradient updates,
compared with 3.08\%, 4.74\%, and 4.74\% for the business baseline.

\textbf{Industrial metrics describe different aspects of ranking and risk retrieval.}
Table~\ref{tab:industrial} summarizes AUROC, Passed-CB AP, and relative
AddCB change from the industrial evaluations.
The TAPES AUROC entry of 92.04\% comes from native recovery;
the Scratch entry of 90.88\% comes from business adaptation.
The corresponding within-setting AUROC comparisons are illustrated in
Figure~\ref{fig:industrial-finetuning}(a,b).
For Passed-CB AP, the table records 5.80\% for Scratch, 5.47\% for
MLEM, and 4.48\% for TAPES, describing precision--recall performance
on the Passed-CB subset.
The AddCB column reports recall changes at a 0.5\% intervention budget
using the reference models associated with the respective evaluation
records. Together, these measures cover population-level ranking,
subset precision--recall behavior, and risk capture at a specified budget.

\subsection{Ablation Studies}
\label{sec:exp-ablation}

To assess the individual contributions of target-set selection and joint
prediction, we conduct ablation studies on AlfaBattle and Retail X5
under three configurations:
(1) removing target-set selection and applying Joint prediction to a fixed set;
(2) replacing Joint with Marginal prediction on the Joint-selected set; and
(3) removing both components, using a fixed set with Marginal prediction.
Together with TAPES, these variants yield four pretrained configurations,
with training from scratch as an additional reference.
The pretrained configurations share the mixture-head parameterization;
Table~\ref{tab:component-ablation} reports the results.

\textbf{The complete configuration has the highest primary scores in both reported ablations.}
Relative to no pretraining, TAPES improves AUROC by 2.13
percentage points on AlfaBattle and Macro-F1 by 0.90 points on Retail X5.
All pretrained variants exceed the corresponding no-pretraining score
on these primary metrics.

\textbf{Target selection and joint prediction each provide additional primary-metric gains.}
With Joint prediction, replacing the fixed target set with the selected
set increases AlfaBattle AUROC and Retail X5 Macro-F1 by 0.29 and 0.68
percentage points, respectively.
On the selected set, Joint improves over Marginal by 0.26 and 0.29 points.
The selected-Marginal control uses the target set chosen under Joint.
Retail X5 AUROC is tied at 79.90 at the displayed precision.

%% file: sections/Conclusion.tex
\section{Conclusion}
\label{sec:conclusion}

We presented TAPES, a task-aligned joint predictive pretraining
approach for transaction event sequences.
TAPES constructs predictive targets through field--operator pairings,
selects target sets according to downstream representation utility,
and learns history representations through a joint likelihood with
a shared discrete latent variable.
Experiments on five public benchmarks demonstrate competitive
performance across prediction tasks and encoder architectures,
while component analyses examine the roles of target-set selection
and joint prediction.
These findings highlight the semantics and organization of predictive
supervision as important design choices for transaction representation
learning.
Future work will explore more efficient candidate-set evaluation and
further characterize when target dependencies benefit downstream
learning across tasks and data conditions.

%% file: sections/Appedix.tex



\section{Additional Experimental Details}
\label{app:experimental-details}

\subsection{Datasets}
\label{app:datasets}
The datasets span binary, multiclass, and multilabel prediction tasks.
We describe each benchmark and the industrial dataset below, including
its predictive target, evaluation measures, and data characteristics.
\paragraph{AlfaBattle.}
AlfaBattle
\footnote{\url{https://boosters.pro/championship/alfabattle2/overview}}
contains transaction histories for credit-default prediction.
The downstream label is binary.
We report AUROC as the primary metric and AP as a supplementary metric.

\paragraph{Retail X5.}
Retail X5
\footnote{\url{https://ods.ai/competitions/x5-retailhero-uplift-modeling/data}}
uses purchase histories for four-class age prediction.
We report Macro-F1 as the primary metric, together with accuracy and
macro one-vs-rest AUROC.

\paragraph{MBD-mini.}
MBD-mini
\footnote{\url{https://huggingface.co/datasets/ai-lab/MBD-mini}}
provides banking event histories for next-month purchase prediction
over four banking products.
The task is multilabel classification.
Macro-AUROC and macro-AP average the respective binary metric over
the four labels.

\paragraph{Rosbank Churn.}
Rosbank Churn
\footnote{\url{https://boosters.pro/championship/rosbank1/}}
uses transaction histories to predict customer churn.
The downstream task is binary classification, evaluated using AUROC and AP.

\paragraph{Taobao.}
Taobao
\footnote{\url{https://tianchi.aliyun.com/dataset/46}}
contains e-commerce behavior histories.
The processed task predicts purchases over the following seven days,
with AUROC as the primary metric and AP reported additionally.

\paragraph{TransRisk.}
TransRisk is a large-scale private industrial dataset from a real-world
transaction risk assessment task.
The training and validation partitions used in our experiments contain
24,698,621 and 4,825,198 transaction-level prediction examples,
respectively, totaling over 29 million examples.
Each example associates a current transaction with its corresponding
user history.
Historical events contain 11 categorical attributes, transaction amounts,
and temporal information, with a history length limit of 300 events.
The full business-adaptation model additionally uses a 652-dimensional
feature vector describing the current transaction.
The downstream task is binary transaction-risk prediction.
We evaluate ranking quality and risk retrieval under constrained
intervention budgets.
Table~\ref{tab:transrisk-statistics} summarizes its experimental data.

\begin{table}[t]
\centering
\small
\caption{TransRisk data used in the experiments.}
\label{tab:transrisk-statistics}
\resizebox{\columnwidth}{!}{%
\begin{tabular}{lr}
\toprule
Item & Value \\
\midrule
Training examples & 24,698,621 \\
Validation examples & 4,825,198 \\
Total examples & 29,523,819 \\
Categorical attributes per historical event & 11 \\
Maximum historical events & 300 \\
Current-transaction feature dimension & 652 \\
\bottomrule
\end{tabular}}
\end{table}

\subsection{Baseline Implementations}
\label{app:baselines}
The baseline implementations cover history aggregation, supervised
sequence learning, and event-level pretraining.
We describe their inputs, learning objectives, and evaluated components below.

\paragraph{Tree ensembles on aggregated features.}
CatBoost~\cite{NEURIPS2018_catboost} and XGBoost~\cite{xgboost2016}
take statistical features aggregated from each event history as input.
These history-level feature vectors summarize behavioral information
for supervised prediction. Both methods are fitted using downstream
training labels and evaluated on the corresponding task.
The main table uses the method names CatBoost and XGBoost for
these aggregated-feature implementations.

\paragraph{Supervised sequence encoders.}
GRU~\cite{cho2014learning}, LSTM~\cite{hochreiter1997lstm},
and causal Transformer~\cite{vaswani2017attention} baselines encode
ordered event histories using recurrent updates or causal attention.
The last valid event state provides the sequence representation
for a task-specific prediction head.
The encoder and head are trained together using downstream labels.
The encoder study uses these supervised models as Scratch references
for their corresponding TAPES configurations.
Encoder dimensions, RoPE components, and training schedules are
specified in Appendix~\ref{app:models-training}.

\paragraph{Self-supervised sequence pretraining.}
We evaluate project-adapted versions of CoLES, TabBERT, MLEM,
and PANTHER-SPRM. Their learning signals cover contrastive,
reconstructive, and combined representation learning.

\emph{CoLES.}
CoLES~\cite{babaev2022coles} constructs different subsequence views
from an entity's history and learns their representations through
contrastive supervision. The objective encourages consistency
across partial observations of the same entity.

\emph{TabBERT.}
TabBERT~\cite{padhi2021tabular} learns by reconstructing masked
field values from the available observations.
This supplies attribute-level supervision for representations of
heterogeneous event histories.

\emph{MLEM.}
MLEM~\cite{moskvoretskii2024mlem} combines generative modeling
and contrastive learning through alignment of their representations.
The adaptation uses both learning signals to obtain representations
for downstream prediction.

\emph{PANTHER.}
PANTHER  
uses the public Sequence Pattern Recognition
Module configuration of PANTHER~\cite{li2025panther}.
The evaluated model consists of the sequential pretraining branch
and its downstream adaptation.


\subsection{Encoder Configurations and Training}
\label{app:models-training}
We use recurrent and attention-based encoders to examine how TAPES
behaves across sequence-modeling architectures.
The following settings describe their shared inputs, architecture choices,
and training schedules.
The backbone study evaluates GRU and unidirectional LSTM with
hidden dimension 512 and two layers; a standard causal Transformer
with hidden dimension 256 and four layers; and RoPE configurations
with either 256 dimensions and four layers or 1,024 dimensions and eight layers.
These backbones use the same event-field encoding scheme and read
the last valid event state as the sequence representation.

The standard Transformer uses learned absolute positional embeddings,
LayerNorm~\cite{ba2016layer}, and GELU feed-forward layers~\cite{hendrycks2016gelu}.
The RoPE configuration uses rotary positional embeddings~\cite{su2024roformer},
RMSNorm~\cite{zhang2019rmsnorm}, query--key normalization,
and SwiGLU feed-forward layers~\cite{shazeer2020glu}.

The base TAPES recipe uses four targets per candidate set, two mixture
components, and rank-16 prediction heads.
Stage~1 evaluates up to eight candidates, with 512 predictive updates
and 128 probe updates per candidate.
The base Stage~2 budget is five complete passes over the predictive-pretraining
partition.
Public downstream adaptation uses at most ten epochs with early stopping.

In Table~\ref{tab:backbone-comparison}, standard-Transformer TAPES uses
ten Stage~2 epochs on AlfaBattle and Retail X5 and sixty on Rosbank Churn.
The remaining TAPES configurations use five epochs.
The table reports configurations selected from completed seeds and training schedules.

\subsection{Public Evaluation Settings}
\label{app:evaluation-details}
The evaluation separates supervision selection, representation learning,
and downstream prediction so that each stage uses the data assigned
to its learning objective.
Predictive pretraining, probe fitting, and target selection use disjoint
subsets of the downstream training pool.
Final supervised adaptation uses the full training pool.
Checkpoint selection uses the early-stopping partition,
and scores are computed on a separate development evaluation partition.
Each main-table and component-table configuration is evaluated through
one training run and its selected checkpoint.

The label-budget study uses 20\%, 40\%, 60\%, 80\%, and 100\%
of the fine-tuning labels. 
Pretrained checkpoints and Stage~I selection are fixed within each benchmark.
Curves report the mean and sample standard deviation across these seeds.

Avg.~Rank in Table~\ref{tab:main-comparison} is computed from the five
designated primary metrics, using average ranks for ties.
Absolute score differences are expressed in percentage points.
Relative changes in Table~\ref{tab:backbone-comparison} are computed as
$100(s_{\mathrm{TAPES}}/s_{\mathrm{supervised}}-1)$.

\begin{table*}[t]
\centering
\small
\caption{Dataset summary.
Public sequence and event counts refer to the full downstream
training sets after eligibility filtering and history truncation.
Event counts exclude padding; validation and test partitions
are not included in these counts.}
\label{tab:datasets}
\begin{tabular}{llrrl}
\toprule
Dataset & Task & \# Train sequences & \# Train events
& Primary metric \\
\midrule
Alpha
& Credit default
& 674,667 & 106,567,621 & AUROC \\
Retail X5
& Age classification
& 222,822 & 25,324,802 & Macro-F1 \\
MBD-mini
& Product purchase, four labels
& 499,930 & 10,917,772 & Macro-AUROC \\
Rosbank Churn
& Churn prediction
& 3,475 & 333,261 & AUROC \\
Taobao
& Future purchase
& 12,837 & 3,220,892 & AUROC \\
\bottomrule
\end{tabular}
\end{table*}

\begin{algorithm}[t]
\caption{TAPES: Task-Aligned Joint Predictive Pretraining}
\label{alg:tapes}
\small
\begin{algorithmic}[1]

\Require Training set $\mathcal D$; fields $\mathcal F$;
operators $\mathcal O$; candidate rule $\mathcal R$;
common initialization $\theta_0$; components $K$;
budgets $B_A,B_B,B_p$; task metric $Q$

\Ensure Selected target set $S^*$ and downstream predictor
$g_{\psi^*}\circ f_{\theta^*}$

\State Split $\mathcal D$ into disjoint
$\mathcal D_{\mathrm{ssl}},
 \mathcal D_{\mathrm{probe}},
 \mathcal D_{\mathrm{sel}}$

\State Construct admissible field--operator targets
$\mathcal A\subseteq\mathcal F\times\mathcal O$

\State $\eta\gets
\Call{FitTransforms}{\mathcal A,\mathcal D_{\mathrm{ssl}}}$

\State Freeze $\eta$ and fix the ordered candidate bank
\Statex \hspace{\algorithmicindent}
$\mathcal C=(S_1,\ldots,S_M)
\gets\Call{BuildBank}{\mathcal A,\mathcal R}$

\State \textbf{Stage 1: Task-aligned target-set selection}

\For{$j=1,\ldots,M$}
    \State $\theta_j\gets
    \Call{JointPretrain}{S_j,\theta_0,B_A}$

    \State Freeze the event input module and encoder
    $f_{\theta_j}$

    \State $\phi_j\gets
    \Call{FitProbe}{
        f_{\theta_j},\mathcal D_{\mathrm{probe}},B_p}$

    \State $\widehat U_j\gets
    Q(g_{\phi_j}\circ f_{\theta_j};
      \mathcal D_{\mathrm{sel}})$
\EndFor

\State $j^*\gets
\min\operatorname*{arg\,max}_{1\leq j\leq M}
\widehat U_j$;
$S^*\gets S_{j^*}$

\State \textbf{Stage 2: Shared-latent joint pretraining}

\State $\theta^{\mathrm{II}}\gets
\Call{JointPretrain}{S^*,\theta_0,B_B}$

\State \textbf{Downstream adaptation}

\State Initialize a new task head $g_\psi$ and optimizer

\State $(\theta^*,\psi^*)\gets
\Call{FineTune}{\theta^{\mathrm{II}},\psi,\mathcal D}$

\State \Return $S^*$ and
$g_{\psi^*}\circ f_{\theta^*}$

\end{algorithmic}
\end{algorithm}

\section{Algorithm}

Algorithm~\ref{alg:tapes} and Algorithm~\ref{alg:tapes-joint} summarize the TAPES pipeline and its shared joint pretraining routine. Stage 1 selects a target set according to downstream utility, while Stage 2 retains the selected target definitions and frozen transformations and retrains the encoder from the common initialization. Both stages optimize the same joint likelihood, with exact marginalization over shared latent components and support for partially observed targets.

\section{Downstream Adaptation and Evaluation}
\label{app:downstream}

\subsection{Transfer Interface and Prediction Readout}

For standard sequence classification, the event input module
and encoder are initialized from the final Stage~2 checkpoint.
The new task head is an affine layer,
$z_i=W_{\psi}h_{i,T_i}+b_{\psi}$,
where $h_{i,T_i}$ is the last valid event state.
Padding positions are excluded from readout.
We apply sigmoid outputs for binary and multi-label prediction
and softmax outputs for multi-class prediction.
The Stage~1 probe and its representation-standardization
statistics are not reused, while the input module retains
its original preprocessing.

For tasks with an established prediction architecture,
we transfer the corresponding pretrained modules while
retaining the task-specific readout and fusion interfaces.

\subsection{Supervised Adaptation Objective}

Standard adaptation jointly updates the transferred modules
and the new task head:
\begin{equation}
\mathcal{L}_{\mathrm{task}}(\theta,\psi)
=
\frac{1}{|\mathcal{D}_{\mathrm{fit}}|}
\sum_{i \in \mathcal{D}_{\mathrm{fit}}}
\ell_{\mathrm{sup}}(z_i,Y_i).
\label{eq:downstream-adaptation}
\end{equation}
The encoder starts from the Stage~2 parameters, whereas
the task head and optimizer are newly initialized.
Binary classification uses binary cross-entropy.
For fully observed multi-label targets, binary cross-entropy
is averaged over labels and then over examples.
Multi-class classification uses cross-entropy.
The joint event-prediction loss is no longer optimized.
Tasks with established prediction architectures use their
corresponding supervised losses.

\subsection{Training Configuration and Evaluation Rules}

Under the standard protocol, the full training set is the
union of the internal event-pretraining, probe-fitting,
and target-selection subsets.
Once the target set is fixed, all these members and their
downstream labels participate in supervised fine-tuning.
Stage~2 event pretraining remains restricted to its original
event-pretraining subset.
Neither the early-stopping validation set nor the development
evaluation set is included in fine-tuning.

Table~\ref{tab:adaptation-settings} summarizes the reference
configuration for standard sequence classification.
The reference encoder learning rates are
$3\times10^{-4}$ for GRU/LSTM,
$10^{-4}$ for the standard causal Transformer and compact
Velocity, and $5\times10^{-5}$ for large Velocity.
Learning rates remain constant during fine-tuning.
Training members are reshuffled and traversed without
replacement in each epoch.

\begin{table}[t]
\centering
\caption{Reference configuration for standard
sequence-classification adaptation.}
\label{tab:adaptation-settings}
\small
\begin{tabular}{
@{}p{0.48\columnwidth}p{0.46\columnwidth}@{}
}
\hline
Setting & Configuration \\
\hline
Optimizer & AdamW \\
Weight decay & $0.01$ \\
Gradient-norm clipping & $1.0$ \\
Effective batch size & At most $128$; final batch retained \\
Task-head learning rate & $10^{-3}$ \\
Maximum/minimum epochs & $10/3$ \\
Early-stopping patience & $3$ epochs \\
\hline
\end{tabular}
\end{table}

At the end of each epoch, we compute the prespecified primary
task metric on a dedicated early-stopping validation set.
The best checkpoint is updated only when the score exceeds
the current best by more than $10^{-12}$; otherwise, the
previous checkpoint is retained.
After the minimum training duration, training stops following
three consecutive epochs without improvement.
The selected checkpoint is subsequently evaluated on a
development set separate from the early-stopping set.
Since this development set participates in configuration
development, its results are not treated as independent
final-test performance.
Optimization and module-freezing settings for task-specific
business prediction architectures are reported separately.

\begin{algorithm}[t]
\caption{Shared-Latent Joint Predictive Pretraining}
\label{alg:tapes-joint}
\small
\begin{algorithmic}[1]

\Require Shared pretraining split $\mathcal D_{\mathrm{ssl}}$,
frozen target transformations $\eta$,
and component count $K$

\Ensure Pretrained event input module and encoder
parameters $\theta$

\Function{JointPretrain}{$S,\theta_0,B$}

    \State $\theta\gets\theta_0$;
    initialize a fresh joint prediction head $\omega$

    \State Reset the optimizer and common
    pretraining sample stream

    \For{$b=1,\ldots,B$}

        \State Draw an effective update batch
        $\mathcal B_b$ from $\mathcal D_{\mathrm{ssl}}$

        \ForAll{$n\in\mathcal B_b$}

            \State $(\mathbf y_n,\mathbf m_n)\gets
            \Phi_S(H_n,e_n^+;\eta_S)$

            \State $h_n\gets f_\theta(H_n)$;
            $O_n\gets\{a\in S:m_{n,a}=1\}$

            \State $\begin{aligned}[t]
            s_{n,k}\gets{}&
            \log\pi_k(h_n)\\
            &+\sum_{a\in O_n}
            \log q_{a,k}(y_{n,a}\mid h_n),
            \quad k=1,\ldots,K
            \end{aligned}$

            \State $\ell_n\gets
            -\operatorname{logsumexp}_{k=1}^{K}s_{n,k}$

        \EndFor

        \State $\mathcal V_b\gets
        \{n\in\mathcal B_b:O_n\neq\varnothing\}$;
        assert $|\mathcal V_b|>0$

        \State $\displaystyle
        \mathcal L_J\gets
        \frac{1}{|S|\,|\mathcal V_b|}
        \sum_{n\in\mathcal V_b}\ell_n$

        \State $(\theta,\omega)\gets
        \Call{OptimizerStep}{
            \theta,\omega,
            \nabla_{\theta,\omega}\mathcal L_J}$

    \EndFor

    \State \Return $\theta$

\EndFunction

\end{algorithmic}
\end{algorithm}

\section{More Experimental Results}
\begin{figure}[htbp]
  \centering
 \includegraphics[width=\linewidth]{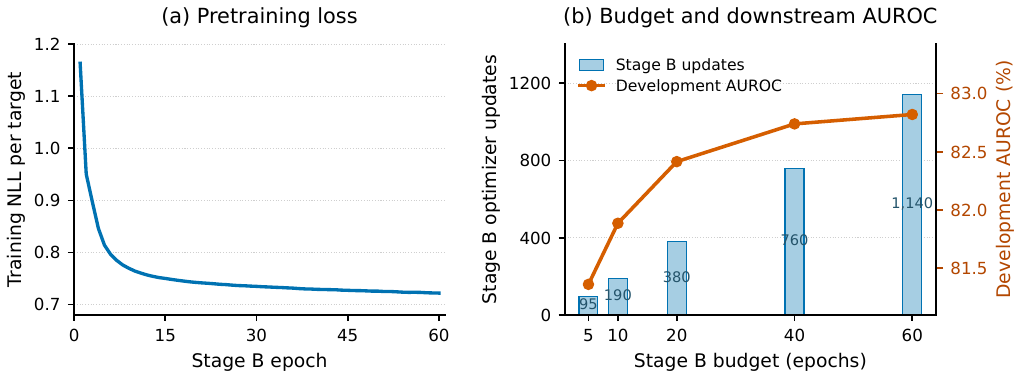}
 \caption{
Effect of the Stage~2 pretraining budget on Rosbank Churn
with the CT256$\times$4 backbone. 
(a) Training negative log-likelihood per target over 60 epochs.
(b) Stage~2 optimizer updates (bars, left axis) and development
AUROC after fine-tuning (line, right axis) at
5, 10, 20, 40, and 60 epochs.
The Stage~1-selected target set, fitted transformations,
and fine-tuning protocol are fixed across budgets.
Each score uses the fine-tuning checkpoint selected by
early-stopping AUROC.
}
\label{fig:pretraining-budget}
\end{figure}

\paragraph{Sensitivity to the pretraining budget.}
We examine the effect of Stage~2 training duration on
Rosbank Churn using the CT256$\times$4 backbone.
We fix the Stage~1-selected target set, fitted transformations,
and fine-tuning protocol, and compare pretraining budgets of
5, 10, 20, 40, and 60 epochs. 
As shown in Figure~\ref{fig:pretraining-budget},
training NLL decreases rapidly during the early epochs
and then declines more slowly.
Increasing the budget from 5 to 60 epochs improves development
AUROC from 81.36\% to 82.82\% (+1.46 percentage points).
However, extending training from 40 to 60 epochs yields only
0.08 additional percentage points, despite a 50\% increase
in Stage~2 updates (760 to 1,140).
These results indicate diminishing returns from additional pretraining on Rosbank Churn. 

